\documentclass[runningheads]{llncs}
\usepackage[T1]{fontenc}
\usepackage{graphicx,verbatim}
\usepackage{graphbox}
\graphicspath{{Imagenes/}}
\usepackage{hyperref}
\usepackage{color}

\usepackage{amsmath}
\usepackage{amssymb}

\usepackage{epsfig}
\usepackage{multirow}
\usepackage{svg}
\usepackage{cancel}
\usepackage{pifont}\newcommand{\xmark}{\ding{55}}

\usepackage[pscoord]{eso-pic}\newcommand{\placetextbox}[3]{\setbox0=\hbox{#3}\AddToShipoutPictureFG*{\put(\LenToUnit{#1\paperwidth},\LenToUnit{#2\paperheight}){\vtop{{\null}\makebox[0pt][c]{\begin{tabular}{l}#3\end{tabular}}}}}}

\newcommand{\repeatthanks}{\textsuperscript{\thefootnote}}

\begin{document}
\placetextbox{0.5}{.96}{\small \textit{This preprint has not undergone peer review or any post-submission improvements or corrections.} \\
\small \textit{The Version of Record of this contribution is published in MICCAI 2025, and is available} \\
\small \textit{online at: https://doi.org/[insert DOI]}}
\title{EndoMetric: Near-Light Monocular Metric Scale Estimation in Endoscopy}
\newcommand\anonymize{0}  \if\anonymize0
\author{Raúl Iranzo\thanks{These authors contributed equally to this work}
\and Víctor M. Batlle\repeatthanks
\and Juan D. Tardós\and José M.M. Montiel}
\authorrunning{R. Iranzo et al.}
\institute{Instituto de Investigaci\'on en Ingenier\'ia de Arag\'on (I3A), Universidad de Zaragoza, 
Mar\'ia de Luna 1, 50018 Zaragoza, Spain.
\email{\{riranzo,vmbatlle,tardos,josemari\}@unizar.es}}

\else

\author{Anonymized Authors}  \authorrunning{Anonymized Author et al.}
\institute{Anonymized Affiliations \\
    \email{email@anonymized.com}}
    
\fi

\maketitle

\begin{abstract}
Geometric reconstruction and SLAM with endoscopic images have advanced significantly in recent years. In most medical fields, monocular endoscopes are employed, and the algorithms used are typically adaptations of those designed for external environments, resulting in 3D reconstructions with an unknown scale factor.
 
For the first time, we propose a method to estimate the real metric scale of a 3D reconstruction from standard monocular endoscopic images without relying on application-specific learned priors. Our fully model-based approach leverages the near-light sources embedded in endoscopes, positioned at a small but nonzero baseline from the camera, in combination with the inverse-square law of light attenuation, to accurately recover the metric scale from scratch. This enables the transformation of any endoscope into a metric device, which is crucial for applications such as measuring polyps, stenosis, or assessing the extent of diseased tissue.


\end{abstract}
\section{Introduction}
\label{section:intro}
\noindent

In current endoscopic procedures, endoscope navigation, localization, and tissue measurement are performed manually. Recent advances in Visual Simultaneous Localization and Mapping (VSLAM) for endoscopy~\cite{lamarca2020defslam,ma2021rnnslam,gomez2024nrslam} offer the promise of live 3D reconstructions, that will enable autonomous or assisted navigation and robotized interventions. Most specialties use just monocular endoscopes to reduce bulk and cost. However, using a moving monocular camera, the absolute scale of the environment is unobservable, and the 3D reconstructions and trajectories obtained have an unknown scale factor. This also introduces scale drift which significantly reduces map accuracy.

\begin{figure}[t!]
    \centering
    \begin{tabular}{p{3.5cm}p{4cm}c}
         \multicolumn{3}{c}{\includegraphics[width=0.99\columnwidth]{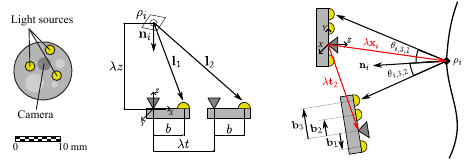}} \\
         (a) Endoscope tip & (b) Simplified model & (c) Full model \\
    \end{tabular}
    \caption{Our method estimates the metric scale factor $\lambda$ by leveraging a near-light illumination model applied to multi-view images captured with a monocular endoscope.}
    \label{fig_esquema_2_camaras_lambda}
\end{figure}

However, endoscopes are equipped with light sources attached to the camera, which introduce significant illumination variations in the scene. Our key insight is to leverage these illumination changes through near-light photometry to accurately recover the true metric scale of monocular reconstructions. Photometry is scale-dependent due to two factors: the inverse-square decay of illumination with distance from the light source, and the angle between the incident light and the surface normal when the light source is positioned at a small, but nonzero, baseline from the camera's optical center (\autoref{fig_esquema_2_camaras_lambda}). 

We demonstrate how true scale can be recovered just from the images captured by a standard monocular endoscope in two steps: first, obtaining an up-to-scale reconstruction with SfM or VSLAM, and then performing photometric optimization to recover scale, gains, and albedo. Our contributions are:
\begin{itemize}
    \item A scale-dependent near-light photometric model applicable to any monocular, up-to-scale multi-view reconstruction.
    \item A photometric optimization method to estimate true metric scale, scene albedo, and camera gain.
    \item An initialization technique to enhance convergence and avoid local minima.
    \item Simulations and real experiments demonstrating the scale accuracy achievable in endoscopy.
\end{itemize}

\section{Related work}
\label{sec:related}

Feature-based monocular SLAM \cite{campos2021orb,elvira2024cuda} and SfM \cite{schonberger2016structure} recover up-to-scale geometry via bundle adjustment, ignoring photometry. 
Photometric monocular SLAM \cite{engel2017direct,zubizarreta2020} estimates geometry, albedo, and camera gains but assumes constant illumination and cannot recover scale. In contrast, we first estimate up-to-scale geometry via bundle adjustment and then determine metric scale, albedo, and camera gain using non-linear optimization of near-light photometric errors.

Our method relates to photometric stereo, originating from \cite{Woodham-80}, which used orthographic images from a fixed camera with distant, switchable light sources to recover surface normals, but not scale. The first method to recover metric scale was \cite{Iwahori-90}, leveraging inverse-square illumination decay from multiple point lights at known positions. Near-field photometric stereo, where the camera and light sources are close to the object, was formalized in \cite{Mecca-14}, with semi-calibrated approaches introduced by \cite{Queau-17}, requiring known light positions but unknown intensities. Using an endoscope for calibrated near-light photometric stereo was first explored in \cite{Collins-13}, combining three colored light sources into a single shot for true-scale reconstruction. In summary, near-light photometric stereo can generate real-scale reconstructions from three or more images with different point light sources. However, for endoscopy, the device must be modified to control the lights or use colored lights, which is impractical in clinical settings.

Other hardware modifications use structured light to overlay a metric scale on the image for polyp measurement \cite{yoshioka2021virtual,vonRenteln2022}. Some works train a deep network to classify polyps into two size classes (smaller or larger that 10~mm) \cite{itoh2018towards,itoh2021binary} or to predict dense depth with true scale for polyp measurement \cite{Du2024polyp,wang2024real}. However, these methods require task-specific learning and do not generalize to other medical applications. In contrast, we achieve metric-scale reconstruction and estimate the camera’s metric trajectory using a standard endoscope without hardware modifications or task-specific learning, relying solely on near-light photometry.  

Illumination decay was previously used in endoscopy in~\cite{batlle2022photometric} to recover 3D information from a single monocular view, assuming constant albedo.  However, their reconstruction remained up-to-scale due to three unknown factors: illumination power, camera gain, and surface albedo. Multi-view near-light photometry to jointly recover scale, gain, and albedo was first proposed in~\cite{Fernandes2022}, but only validated through simplistic simulations with up-to-scale ground-truth geometry. In this work, we take a step further by demonstrating that the method can work from the near-light images alone, both in realistic simulations and in real colonoscopies.

 \section{Fundamentals}

The simplified two-view near-light photometric problem~\cite{Fernandes2022}, is depicted in \autoref{fig_esquema_2_camaras_lambda}b. It assumes a moving monocular camera with a single point light source at a distance $b$ from the optical center, observing a Lambertian point with albedo $\rho$, which lies along the camera's optical axis at a depth $\lambda z$, with its normal pointing towards the camera. The second camera is translated $\lambda t$ to the right. We aim to compute the unknown scale $\lambda$.

Assuming the light intensity $L_0$ and camera gain $g$ are constant, and no gamma compression, the intensity of the point in each image $k$ will be~\cite{Iwahori-90}:
\begin{equation}
    I_k(\lambda) = \frac{\rho g L_0}{\pi}  \frac{\mathbf{n} \cdot \mathbf{l}_k}{ \left\| \mathbf{l}_k \right\|^3 }  
    = \frac{\rho'}{\pi} \frac{\mathbf{n} \cdot \mathbf{l}_k}{ \left\| \mathbf{l}_k \right\|^3 }
\end{equation}
\noindent where $\rho'=\rho g L_0$ is a scaled albedo. The intensities in both images are:
\begin{equation}
    I_1 = \frac{\rho'}{\pi} \frac{\lambda z}{ \left( b^2 + \lambda^2z^2 \right)^{3/2}} \,\,\, , \,\,\, 
    I_2 = \frac{\rho'}{\pi} \frac{\lambda z}{ \left( \left( \lambda t + b  \right)^2 + \lambda^2z^2 \right)^{3/2}}
\end{equation}

We can eliminate $\rho'$ to get a second-order equation on $\lambda$:
\begin{equation}
b^2 + \lambda^2 z^2 = c \left( \left( \lambda t + b \right) ^2 + \lambda^2 z^2  \right)
\end{equation}
where $c=\left( I_2 / I_1 \right)^{2/3}$ is a known constant determined form the intensities measured in the images. This equation allows to obtain $\lambda$, except when $b = 0$, in which case $\lambda$ simplifies away, and cannot be solved. 
 
We can conclude that in a multi-view near-light scenario, metric scale is observable, if the baseline between camera and light source is non-zero, even if light intensity,  gain and albedo are unknown. Also, we can expect scale accuracy to degrade when the distance to the surface is too large compared with the camera-light baseline. The remainder of the paper proposes a practical method for obtaining the scale in real endoscopic settings and studies its accuracy.

\section{EndoMetric}
Our proposed method, EndoMetric, works in two steps: first obtain an \textit{up-to-scale multi-view reconstruction} of the scene using any SLAM or SfM method, and then solve an optimization problem for \textit{metric scale estimation} that impose albedo consistency. The key for scale estimation is leveraging a \textit{near-light photometric model} that takes into account the baseline of the light sources with respect to the camera and the inverse-square law of illumination decline with distance.  

\subsection{Up-to-scale Multi-view Reconstruction}
\label{subsec:geometric}

Classical multiview geometry can produce, from a sequence of calibrated monocular images (at least two), the geometry of $n$ scene points $\left\{\mathbf{x}_i\right\}_{i = 1}^{n}$ and $m$ camera poses $\left\{\mathbf{T}_k=( \mathbf{R}_k, \mathbf{t}_k)\right\}_{k =1}^{m}$ up to an unknown scale factor $\lambda$, by solving bundle adjustment, i.e. the non-linear optimization of the re-projection errors of the points matched along the sequence. We use the well known COLMAP~\cite{schonberger2016structure} to compute multiview geometry from endoscopic sequences.
Photometric models need not only the sparse scene geometry, but also the surface normals at each scene point. We propose to compute the normals $\mathbf{n}_{i}$ by fitting a plane to the $p$ neighbors of each scene point.

\subsection{Near-light Photometric Model}
\label{subsec:photometric}

We assume a calibrated mobile camera with $r$ fixed point light sources at known positions relative to the optical center $\{\mathbf{b}_j\}_{j=1}^{r}$. All lights share the same intensity $L_0$, uniformly distributed in all directions, with a calibrated lens vignetting. Points near specular reflections are discarded, and the surface is assumed to be Lambertian with an unknown, varying albedo. 

We have $m$ grayscale images  $\left\{I_k\right\}_{k = 1}^{m}$ taken from $m$ poses while observing $n$ scene points. For a given point $i$ in an image $k$ the image intensity depends on the point albedo $\rho_i$, the camera gain $g_k$ and the light intensity $L_0$. The observed albedos are always multiplied by the camera gain, $g_k$, and $L_0$. As these values are not provided by currently available endoscopes, we define two new variables that are observable from the images: 
\begin{equation} \label{eq_gain}
    \rho'_i = \rho_i g_1 L_0\;\;\; , \;\;\;g'_k = \frac{g_k}{g_1}
\end{equation}
where $\rho'_i$ is a scaled albedo (that may be greater than 1) and $g'_k$ represents the gain change with respect to the first image.

Illumination depends on the incidence angle and the inverse-square of the distance between light and point. Then, the perceived radiance depends on the unknown scale factor $\lambda$ of the multi-view reconstruction as (\autoref{fig_esquema_2_camaras_lambda}c):
\begin{equation} \label{eq_I_scale_final}
    \mathcal{I}_{i,k}(\rho'_i, g'_k, \lambda) = 
      \left( \frac{\rho'_i g'_k}{\pi} \sum_{j=1}^{r} \frac{\cos \theta_{i,j,k}(\lambda)}{\left\| \lambda \mathbf{x}_i - (\mathbf{R}_{k}\ \mathbf{b}_j + \lambda \mathbf{t}_{k}) \right\|^2} V(\mathbf{x}_i) \right)   ^{1 / \gamma}
\end{equation}
where $  \cos \theta_{i,j,k} (\lambda) = \mathbf{n}_{i}  \frac{\lambda\mathbf{x}_i - (\mathbf{R}_{k}\ \mathbf{b}_j + \lambda\mathbf{t}_{k})}{\left\| \lambda\mathbf{x}_i - (\mathbf{R}_{k}\ \mathbf{b}_j + \lambda\mathbf{t}_{k}) \right\|} $, $V(\mathbf{x}_i)$ is the calibrated lens vignetting and $\gamma$ is gamma compression.

\subsection{Metric Scale Estimation}
Given the up-to-scale reconstruction, our near-light photometric model, and the original images $I_1 ...I_{m}$, our goal is to recover the scale factor $\lambda$ and, as a side product, the point albedos $\rho'_i$ and camera gain changes $g'_k$. This can be achieved minimizing the photometric error with respect to the model:  

\begin{equation} \label{eq_minimizar}
    \underset{\left\{\lambda, \rho'_1 ... \rho'_{n}, g'_2 .. g'_{m}\right\}}{\arg\min}\sum_{i,k} \lVert \mathcal{I}_{i,k}(\rho'_i, g'_k, \lambda) - I_{i,k} \rVert_\epsilon ^2
\end{equation}

\noindent where $I_{i,k}$ is the intensity of point $i$ observed in image $I_k$. A robust cost function is used to reduce the influence of spurious data. This nonlinear optimization is solved using  the Levenberg-Marquardt method implemented in Ceres~\cite{Ceres}.

\subsection{Initial Guess for the Scale}

To avoid local minima and achieve faster convergence it is crucial to find good initial values for the optimization variables ($\lambda$, $\rho'_i$, and $g'_k$). Indeed, these variables are closely related, therefore, instead of estimating their initial values separately as in~\cite{Fernandes2022}, we propose to estimate them jointly. 

We perform an exhaustive search for the scale parameter $\lambda$ over a logarithmic space $\Lambda$. For each trial value $\hat{\lambda}$, we estimate the albedo values $\hat{\rho}'_i$  solving \eqref{eq_I_scale_final} for each point in the first image. Then, to estimate the relative gains of the rest of images, we perform a robust regression. First, we undo the gamma compression $I^\gamma$ to work in linear space. Next, we find the gain value $\hat{g}_k$ that minimizes the difference between the real value $I_{i,k}^\gamma$ and the estimated value $\mathcal{I}_{i,k}^\gamma$ of the points using the robust cost function. Finally, we select the value of $\hat{\lambda}$ with the lowest residual according to \eqref{eq_minimizar}.

\section{Experiments}

\subsection{Datasets}
\label{subsec:dataset}

\paragraph{Simulation dataset.} Real endoscopic images with a ground-truth metric scale are difficult to obtain. Therefore, we assessed the accuracy of our method through simulations. Our endoscope has a monocular fisheye camera and three surrounding light sources with a ${\sim}3$mm baseline (\autoref{fig_esquema_2_camaras_lambda}a). We used a real 3D mesh from \cite{incetan2021vrcaps}, a Lambertian illumination model, and Gaussian pixel noise of 4 gray levels. Examples are available in the supplementary video.

\paragraph{EndoMapper dataset~\cite{azagra2023endomapper}.} We validate our method using real endoscopy videos to assess its performance in real-world conditions. To our knowledge, EndoMapper is the only dataset that provides both photometric calibration of the endoscope and metric-scale annotations of polyp sizes estimated by endoscopists.

\subsection{Impact of Distance to the Surface}

\begin{figure}[t]
    \centering
    \includegraphics[width=0.99\columnwidth]{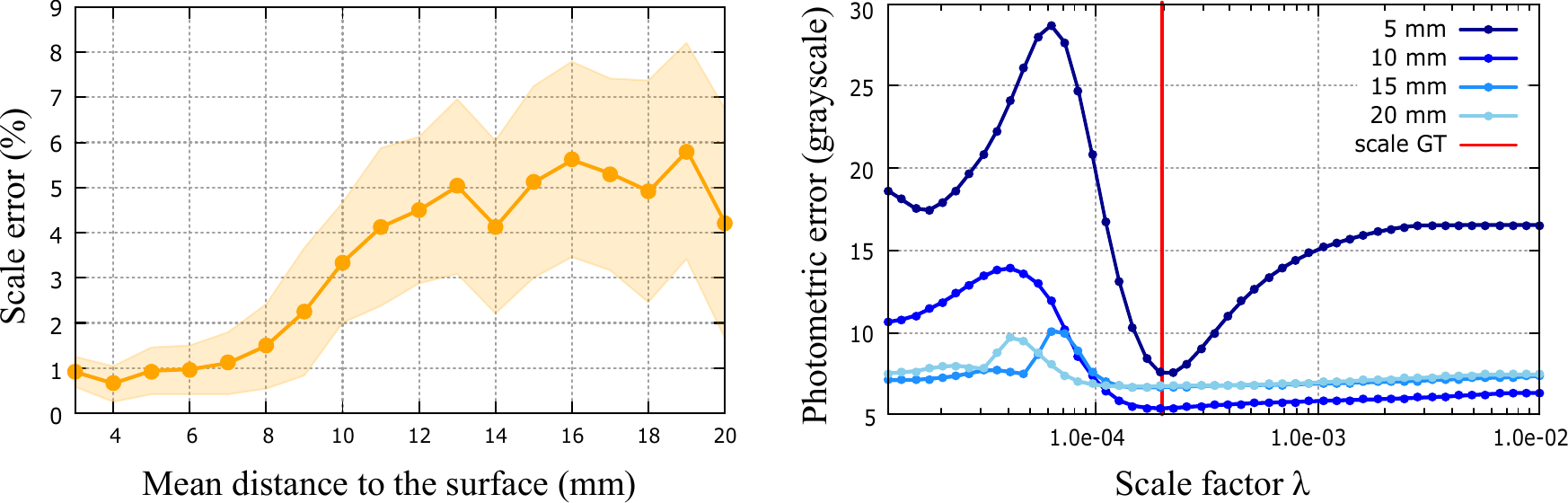}
    \caption{Method accuracy depends on surface distance, performing best when it matches the light-camera baseline. \textit{Left:} Scale error increases from 1\% to 5\% with distance. \textit{Right:} Greater distances weaken the photometric cost function’s minimum.}
    \label{fig:scale_estim_distance}
\end{figure}

In our \textit{simulation dataset}, the endoscope is positioned to face a polyp at varying distances from the surface (3 to 20 mm), capturing images from slightly different viewpoints, which could be easily achieved in practice by bending the endoscope tip. Four images are used at a time to reconstruct the 3D shape with COLMAP and estimate its scale. \autoref{fig:scale_estim_distance}L presents the average scale error as a function of distance to the surface. Since different image sets may yield varying results, we report the mean and standard deviation over five experiments.

The method achieves an error of approximately 1\% at distances up to 8~mm, which are typical in endoscopic procedures. This confirms its effectiveness in near-field conditions, where the distances between the camera, light sources, and surface are of the same order of magnitude. The scale error raises to 5\% as we move away from the surface, since the photometric cost function exhibits a less marked minimum (\autoref{fig:scale_estim_distance}R).

\subsection{Impact of Multi-View Reconstruction Accuracy}
\label{subsec:accuracy}

\begin{table}[t]
\caption{\label{tab:experiments} Accuracy at near-field ranges (5mm). The study highlights the importance of precise multi-view geometry $\mathbf{T}_{k}$ and the benefits of our initial guess.}
\centering
\begin{tabular}{|cc|ccc|c|c|c||ccc|}
\hline
& & \multicolumn{3}{c|}{Optimized} & & & & \multicolumn{3}{c|}{\% error} \\
\multirow{-2}{*}{} & & $\;\;\lambda\;\;$ & $\;\;\rho'\;\;$ & $g'$ & \multirow{-2}{*}{\begin{tabular}[c]{@{}c@{}}Initial \\ guess\end{tabular}} & \multirow{-2}{*}{$\mathbf{T}_{k}$} & \multirow{-2}{*}{$\mathbf{n}_i$} & $\lambda$ & $\rho'$ & $g'$ \\ \hline \hline
 \multicolumn{2}{|c|}{EndoMetric} & \multicolumn{1}{c|}{$\checkmark$} & \multicolumn{1}{c|}{$\checkmark$} & $\checkmark$ & $\checkmark$ & SfM & SfM & \multicolumn{1}{c|}{0.95} & \multicolumn{1}{c|}{5.48} & 3.37 \\ \hline \hline
 & A & \multicolumn{1}{c|}{$\checkmark$} & \multicolumn{1}{c|}{$\checkmark$} & $\checkmark$ & $\checkmark$ & SfM & \textbf{GT} & \multicolumn{1}{c|}{0.81} & \multicolumn{1}{c|}{4.52} & 4.04 \\
 & B & \multicolumn{1}{c|}{$\checkmark$} & \multicolumn{1}{c|}{$\checkmark$} & \textbf{GT} & $\checkmark$ & SfM & GT & \multicolumn{1}{c|}{0.80} & \multicolumn{1}{c|}{3.99} & -- \\
 & C & \multicolumn{1}{c|}{$\checkmark$} & \multicolumn{1}{c|}{$\checkmark$} & GT & $\checkmark$ & \textbf{GT} & GT & \multicolumn{1}{c|}{0.17} & \multicolumn{1}{c|}{3.44} & -- \\ \cline{3-11}
\noalign{\vskip-6\tabcolsep \vskip-3\arrayrulewidth \vskip\doublerulesep}
\\ \cline{3-11}
 & D & \multicolumn{1}{c|}{$\checkmark$} & \multicolumn{1}{c|}{$\checkmark$} & $\checkmark$ & \xmark & SfM & SfM & \multicolumn{1}{c|}{38.21} & \multicolumn{1}{c|}{18.43} & 3.88 \\
 \multirow{-5}{*}{\begin{tabular}[c]{@{}c@{}}Ablation \\ study \end{tabular}} & E & \multicolumn{1}{c|}{$\checkmark$} & \multicolumn{1}{c|}{$\checkmark$} & $\checkmark$ & \cite{Fernandes2022} & GT & GT & \multicolumn{1}{c|}{1.23} & \multicolumn{1}{c|}{4.25} & 4.57 \\ \hline
\end{tabular}

\end{table} 
Focusing on near-field conditions, \autoref{tab:experiments} presents an ablation study using only images from our \textit{simulation dataset} captured at approximately 5~mm from the surface. Results show that scale accuracy remains similar whether ground-truth surface normals (row A) or camera gain values (row B) are provided. However, the error decreases significantly in the scenario with the most ground-truth information (row C), suggesting that improving the underlying multi-view reconstruction could enhance the accuracy of our scale estimation method.

\subsection{Impact of Initial Guess}

Our photometric cost function is non-linear and can exhibit multiple local minima, as illustrated in \autoref{fig:scale_estim_distance}R.
To ensure robustness and generality, we propose a method for computing the initial seed for non-linear optimization. Without this step, our method may get trapped in a sub-optimal solution, resulting in large errors, as shown in \autoref{tab:experiments} (row D). Previous work~\cite{Fernandes2022} initialized albedo and gains to constant values and assumed known camera poses and scene geometry. We tested this configuration with our software and achieved an error of 1.23\% in estimating the real scale under these ideal conditions (row E). In contrast, despite estimating camera poses and scene geometry automatically, our method obtains a smaller error of 0.95\%, thanks to a better initial guess.

\subsection{Real Polyps Measurement}
\label{subsec:real-polyps}

In the \textit{EndoMapper dataset}, some sequences include metadata with annotations derived from the endoscopist’s speech recorded during the procedure. We selected five polyps where the practitioner estimated the lesion size. The selection criteria included a clear view of the polyp, a distance to the surface of 5–15mm, and smooth endoscope tip motion.

\begin{figure}[t]
\begin{tabular}{cccccc}
     & Seq\_041 & Seq\_34 & Seq\_058 & Seq\_041 & Seq\_022 \\
     & Polyp A & Polyp A & Polyp A & Polyp D & Polyp A \\
     Image & \includegraphics[align=c,width=0.155\columnwidth]{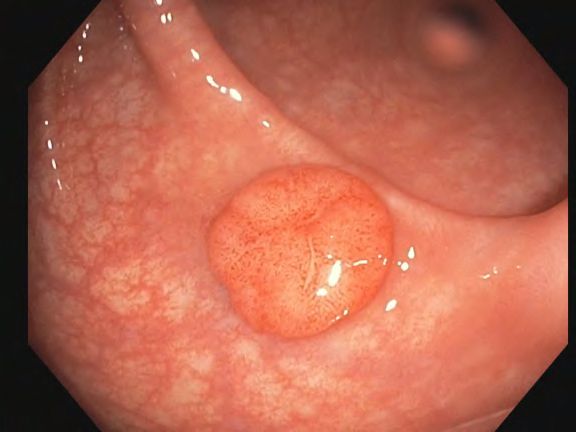} & \includegraphics[align=c,width=0.155\columnwidth]{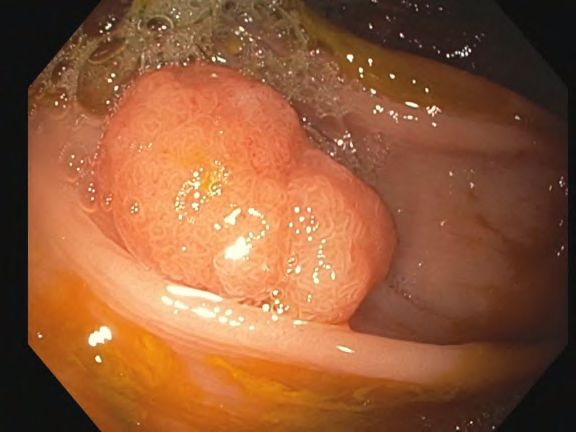} & \includegraphics[align=c,width=0.155\columnwidth]{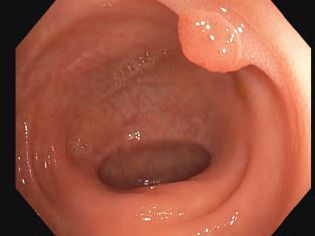} & \includegraphics[align=c,width=0.155\columnwidth]{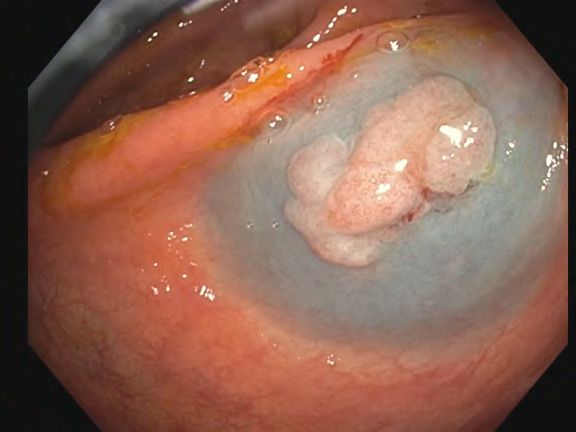} & \includegraphics[align=c,width=0.155\columnwidth]{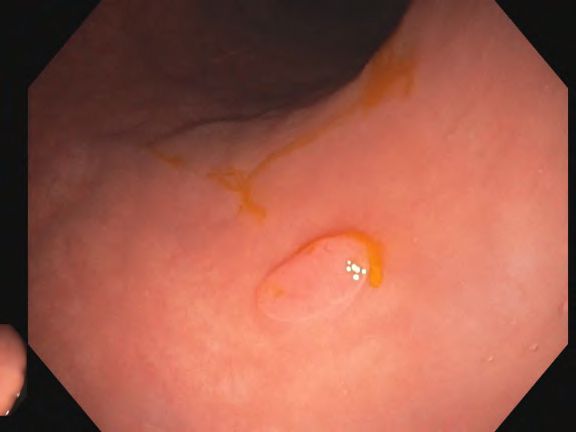} \\
     3D & \includegraphics[align=c,width=0.11625\columnwidth]{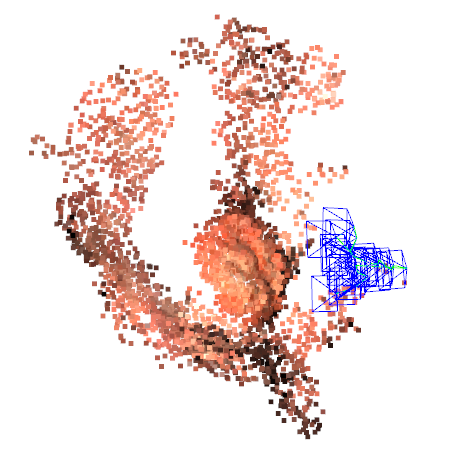} & \includegraphics[align=c,width=0.11625\columnwidth]{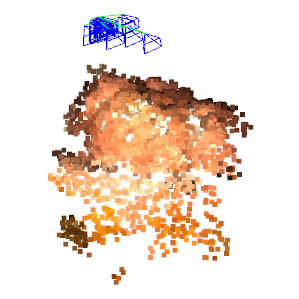} & \includegraphics[align=c,width=0.11625\columnwidth]{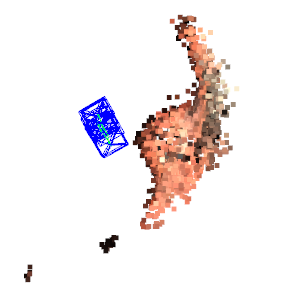} & \includegraphics[align=c,width=0.11625\columnwidth]{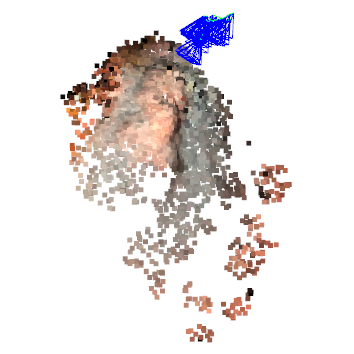} & \includegraphics[align=c,width=0.11625\columnwidth]{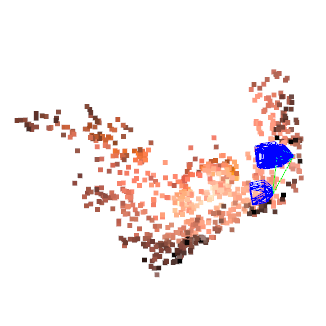} \\
     Normals & \includegraphics[align=c,width=0.11625\columnwidth]{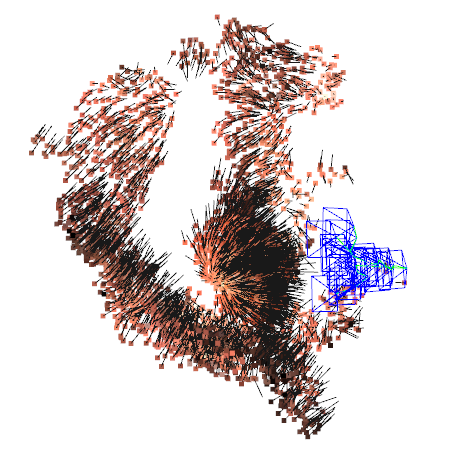} & \includegraphics[align=c,width=0.11625\columnwidth]{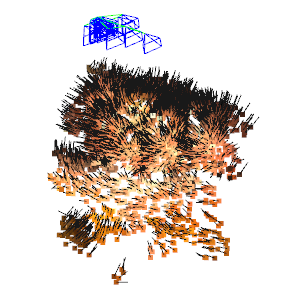} & \includegraphics[align=c,width=0.11625\columnwidth]{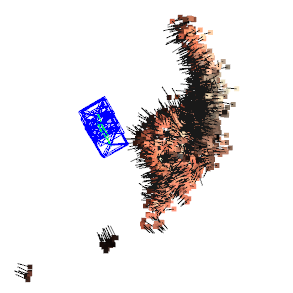} & \includegraphics[align=c,width=0.11625\columnwidth]{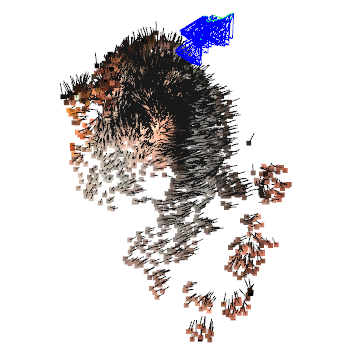} & \includegraphics[align=c,width=0.11625\columnwidth]{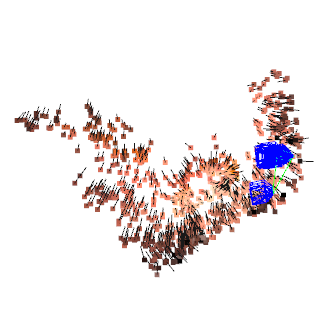} \\
     Result & \includegraphics[align=c,width=0.155\columnwidth]{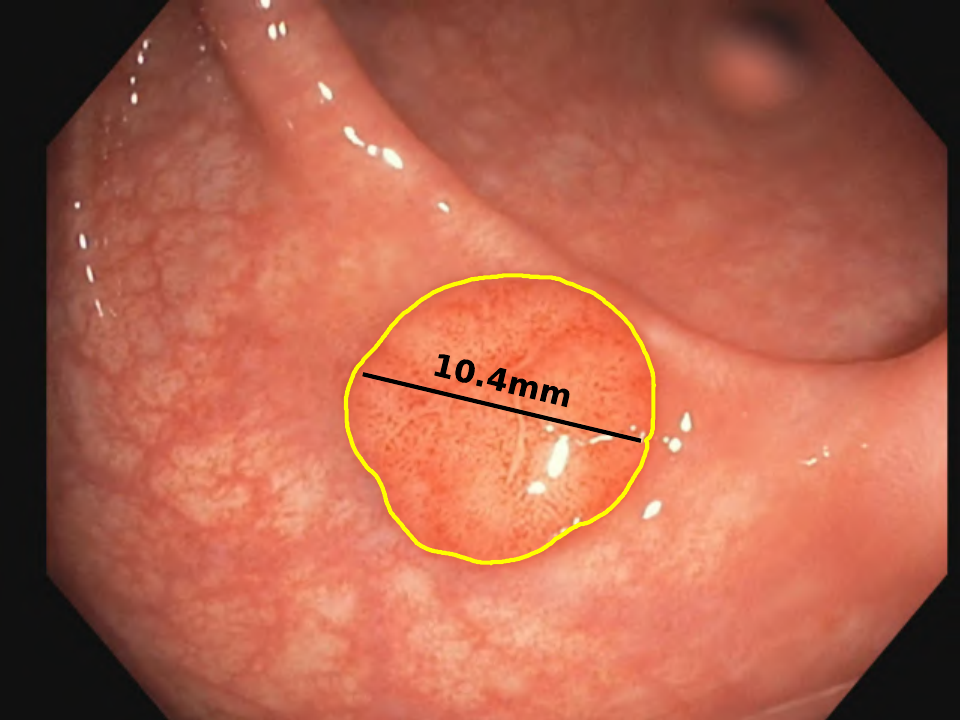} & \includegraphics[align=c,width=0.155\columnwidth]{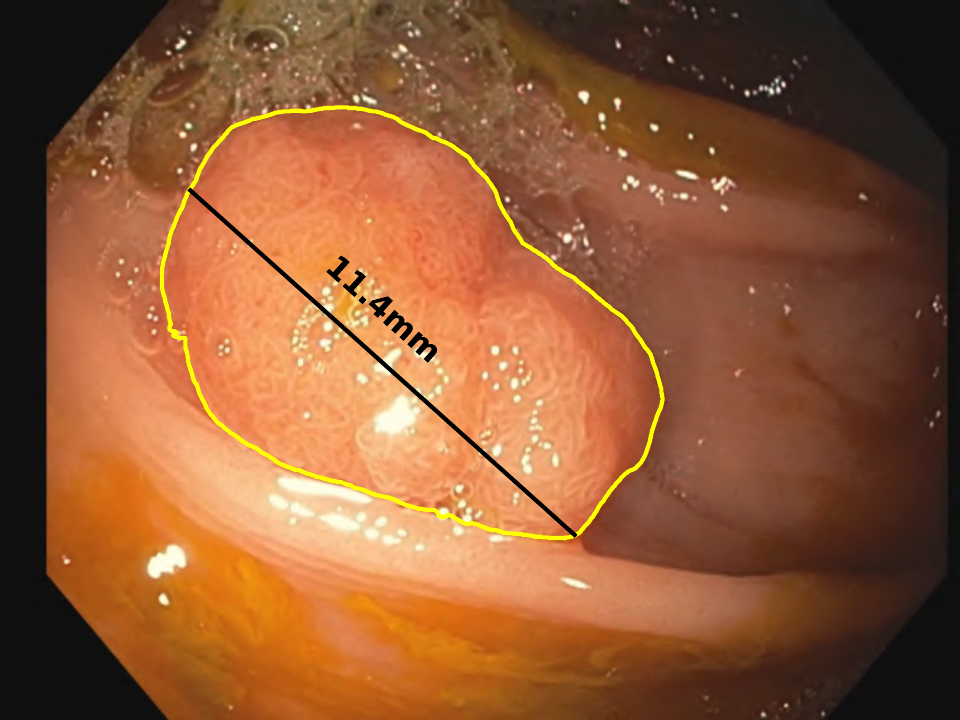} & \includegraphics[align=c,width=0.155\columnwidth]{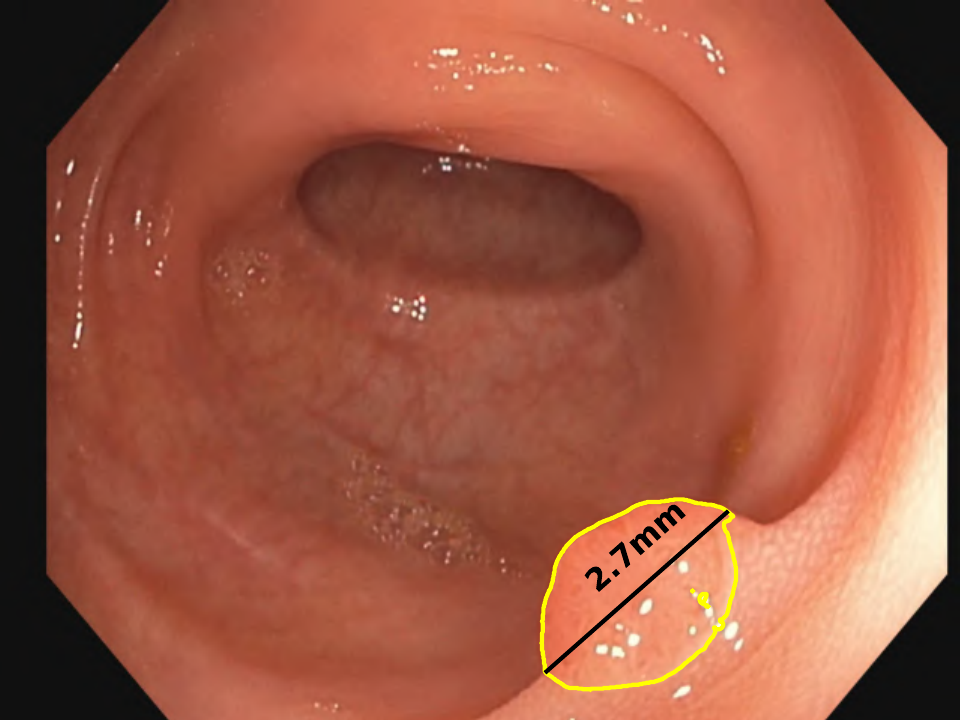} & \includegraphics[align=c,width=0.155\columnwidth]{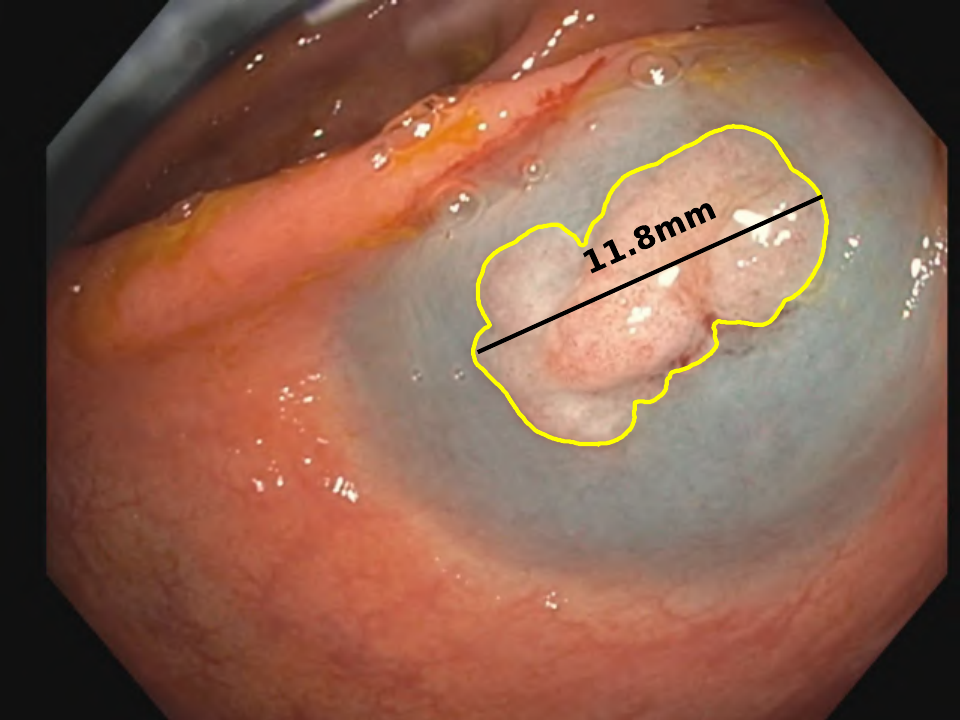} & \includegraphics[align=c,width=0.155\columnwidth]{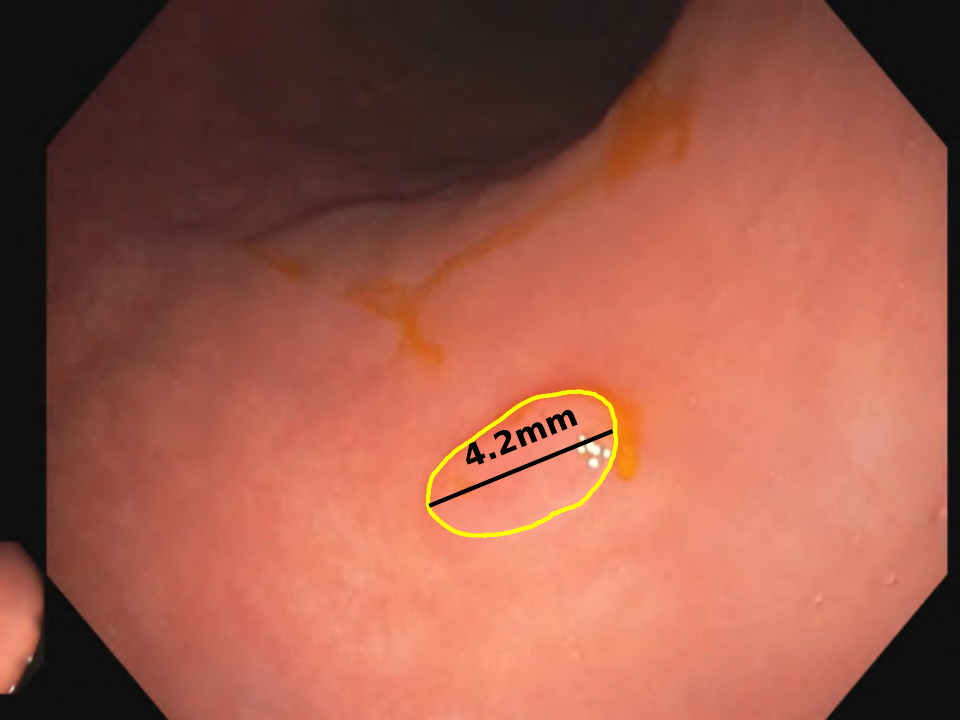} \\
     Our method & 10.4mm & 11.4mm & 2.7mm & 11.8mm & 4.2mm \\
     Endoscopist & 10mm & 10mm & 3mm & 8--10mm & 4--5mm \\
     Discrepancy & 0.4mm & 1.4mm & 0.3mm & 2.8mm & 0.3mm \\
     & (4\%) & (14\%) & (10\%) & (31\%) & (7\%)
     
\end{tabular}
\caption{Results in EndoMapper dataset~\cite{azagra2023endomapper}.}
\label{fig:real-polyps}
\end{figure}

In this real-world setting, we use approximately 20 frames with COLMAP to reconstruct the 3D shape and estimate its metric scale. We then apply Segment Anything~\cite{kirillov2023segment} to identify the polyp within a region of interest in a single frame. The polyp size is determined by measuring the longest diameter of the 3D points within the segmented region. \autoref{fig:real-polyps} presents an input image, the reconstructed 3D shape, the estimated surface normals, and our diameter measurement within the polyp boundaries. On average, our measurements deviate from the endoscopist’s estimation by 1.0 mm (13\%), demonstrating the method's potential for standardizing polyp size assessment.

\section{Conclusions}

We have presented, for the first time, a method to obtain 3D reconstructions with real metric scale from a conventional monocular endoscope, solely based on physical principles. The method does not require any application-specific learning, prior knowledge, or hardware modifications — only a calibrated light-camera setup. 
Our simulations demonstrate that accurate metric scale recovery is achievable in practical conditions. Our experiments on the EndoMapper dataset show that the method produces polyp measurements closely matching those estimated visually by an endoscopist. This provides a preliminary yet solid proof that the proposed method bridges the simulation-to-real gap and effectively estimates metric scale in real images. A quantitative evaluation of its accuracy will require a dataset with ground-truth polyp size annotations.

Exploiting near-field illumination paves the way for real-scale visual SLAM with standard monocular endoscopes. This will be critical in the short term for accurate measurements, and in the long term, for autonomous robotic exploration and surgery.

\if\anonymize0

\begin{credits}
\subsubsection{\ackname} This work was supported by the EU-H2020 grant 863146: ENDOMAPPER, Next Generation EU \textit{Programa Investigo -116-69}, the Spanish government grants PID2021-127685NB-I00 and FPU20/06782, and by Arag\'on government grant DGA\_T45-17R.

\subsubsection{\discintname}
The authors have no competing interests to declare that are
relevant to the content of this article.
\end{credits}

\fi

\bibliographystyle{splncs04}
\bibliography{bibliography}

\end{document}